\documentclass[runningheads]{llncs}
\usepackage{graphicx}
\usepackage{comment}
\usepackage{amsmath,amssymb} % define this before the line numbering.
\usepackage{color}

\usepackage[width=122mm,left=12mm,paperwidth=146mm,height=193mm,top=12mm,paperheight=217mm]{geometry}
\usepackage{soul}
\usepackage{url}
\usepackage[utf8]{inputenc}
\usepackage{booktabs}
\usepackage{subfigure}
\usepackage[ruled,vlined]{algorithm2e}
\usepackage{multirow}
\usepackage{bbding}

\begin{document}

\pagestyle{headings}
\mainmatter
\def\ECCVSubNumber{5094}  % Insert your submission number here

\title{Refinements in Motion and Appearance for Online Multi-Object Tracking}

% CAMERA READY SUBMISSION
%\begin{comment}
\titlerunning{MIFT}

\author{Piao Huang\inst{1} \and
Shoudong Han\inst{1 (}\Envelope\inst{)} \and
Jun Zhao\inst{2} \and
Donghaisheng Liu\inst{1} \and
HongweiWang\inst{1} \and
En Yu\inst{1} \and
Alex ChiChung Kot\inst{2}}
\authorrunning{Huang et al.}

\institute{National Key Laboratory of Science and Technology on Multi-spectral Information Processing, School of Artificial Intelligence and Automation,\\
Huazhong University of Science and Technology,Wuhan,430074,China \and
Nanyang Technological University \\
\email{\{huangpiao,shoudonghan,donghaisheng,hongweiwang,yuen\}@hust.edu.cn, \\
\{junzhao,eackot\}@ntu.edu.sg} }

%******************
\maketitle

\begin{abstract}
Modern multi-object tracking (MOT) system usually involves separated modules, such as motion model for location and appearance model for data association. However, the compatible problems within both motion and appearance models are always ignored. In this paper, a general architecture named as MIF is presented by seamlessly blending the \textbf{M}otion integration, three-dimensional(3D) \textbf{I}ntegral image and adaptive appearance feature \textbf{F}usion. Since the uncertain pedestrian and camera motions are usually handled separately, the integrated motion model is designed using our defined intension of camera motion. Specifically, a 3D integral image based spatial blocking method is presented to efficiently cut useless connections between trajectories and candidates with spatial constraints. Then the appearance model and visibility prediction are jointly built. Considering scale, pose and visibility, the appearance features are adaptively fused to overcome the feature misalignment problem. Our MIF based tracker (MIFT) achieves the state-of-the-art accuracy with 60.1 MOTA on both MOT16\&17 challenges.
\end{abstract}

\section{Introduction}

Multi-object tracking (MOT) plays a crucial role in scene understanding tasks for video analysis. It aims to estimate trajectories of objects and associate them with the given detection results in either online or offline batch way. With recent progress on object detection task, the tracking-by-detection strategy becomes the preferred paradigm to solve the problem of tracking multiple objects. However, despite the advantages of the dependence on detection, it also becomes a major limitation in complex scenes due to the quality of detections.

With the tracking-by-detection paradigm, the tracking task is usually divided into several separate parts, such as motion model, feature extraction, data association. In this paper, we explore the refinements in motion and appearance for online multi-object tracking. The motion model is used to tackle the estimation of pedestrian motion and camera motion, which can be useful for data association. Moreover, the predicted positions by motion models can also be regarded as the trajectories when there exists missed detections. However, the pedestrian motion models and camera motion models are always used separately~\cite{wojke2017simple} or simply combined with each other, the objects' motion states cannot be estimated precisely. To build a motion integration model with high robustness, we explore the relation between non-rigid and rigid motion and integrate them together.

Besides, considering the useless connections between trajectories and candidates, the spatial constrains are applied to each trajectory by 3D integral image during the data association stage. Thus all the detections will be transferred into a feature map in one-hot encoding style (See Section~\ref{integral blocking}). Hence, detections within the target region of each tracking position will be obtained integrally in  constant time complexity, which can significantly decrease the time costs of data association.

Due to the intra-category occlusions and unreliable detections, the extracted features are usually affected by the foreground objects, which are not the targets in the boxes. Each trajectory contains various historical features in different scale, pose and quality, which raises feature misalignments. To resolve this problem, an occlusion aware appearance model is designed to better extract the features of objects with scale invariance and visibility estimations. Considering the misalignments of features, the differences between trajectories' historical features and each coming detection are measured in terms of occlusion (visibility), scale, pose and time gap. Thus the historical features are adaptively fused.

The contributions of our work are summarized as follows:
\begin{itemize}
\item We propose a general architecture (MIF) that can be applied to both of the multi-object tracking and detection tasks with the-state-of-art performance on all MOT benchmarks.
\item We integrate the pedestrian and camera motions to overcome their interaction effects using our proposed motion intension metric.
\item We apply spatial constraints to reduce the time costs of data association by 3D integral image.
\item We design an occlusion aware appearance model and adaptive appearance feature fusion mechanism to handle the misalignments between trajectories and detections.

\end{itemize}

Our code will be released upon acceptance of the paper.

\section{Related Work}

Recent related studies on MOT can be categorized as follows:

\textbf{Motion models for trajectory prediction}.
The motions in video sequences can be summarized as non-rigid motion(pedestrians) and rigid motion(camera pose). The non-rigid motion is commonly described by constant velocity model~\cite{choi2010multiple}. In ~\cite{yang2011learning}, trajectories are smoothed by Gaussian Distribution based on observation. Recently the Kalman Filter tends to be more accepted using the provided detections as observations~\cite{bewley2016simple,wojke2017simple}. Moreover, the social force models are applied due to complex pedestrian motion in crowded scenarios ~\cite{leal2014learning}. As for the rigid motion caused by camera pose variances, researchers have studied in two directions. One is 3D information based methods, such as Ego-motion~\cite{wang2019exploit} and SFM~\cite{choi2010multiple}. The other one is based on affine transformation~\cite{Bergmann_2019_ICCV}. Besides, the conditional probability model with recurrent neural network~\cite{fang2018recurrent} was also proposed to predict the target's position and shape in the next frame. Moreover, the single object tracking(SOT) based methods~\cite{Chu_2019_ICCV,feng2019multi,zhu2018online} were gradually adopted to search targets directly.

\textbf{Appearance feature extraction and selection}.
The identification of target objects and misalignments between trajectories and candidates are the key aspects of appearance feature model. The identification task is commonly regarded as a person re-identification problem~\cite{wojke2017simple}. Due to the affects of background objects and occlusions, the extracted features are usually noisy. To tackle the problem, some spatial attention based methods are adopted to focus on foreground targets~\cite{chen2019aggregate}. As for the various historical features in each trajectory, the misalignments are also needed to be handled by feature selection and fusion tasks. The most direct way is to compare each historical feature with every coming candidate~\cite{wojke2017simple}. However, this kind of method will cost much time with little effects on misalignment problems. To handle this problem, ~\cite{feng2019multi} proposes a quality evaluation model for appearance to select the most representative feature within every time window. In addition, ~\cite{yang2016temporal} explores the temporal dynamic to predict appearance features using the Hidden Markov Model. Recently, the appearance and motion features are also studied to be fused in an end-to-end way, such as the combination of appearance feature with single object tracker (SOT)~\cite{Chu_2019_ICCV} and the joint learning of appearance and location features~\cite{wang2019exploit}.

\begin{figure*}[t]
  \centering
  \includegraphics[width=120mm]{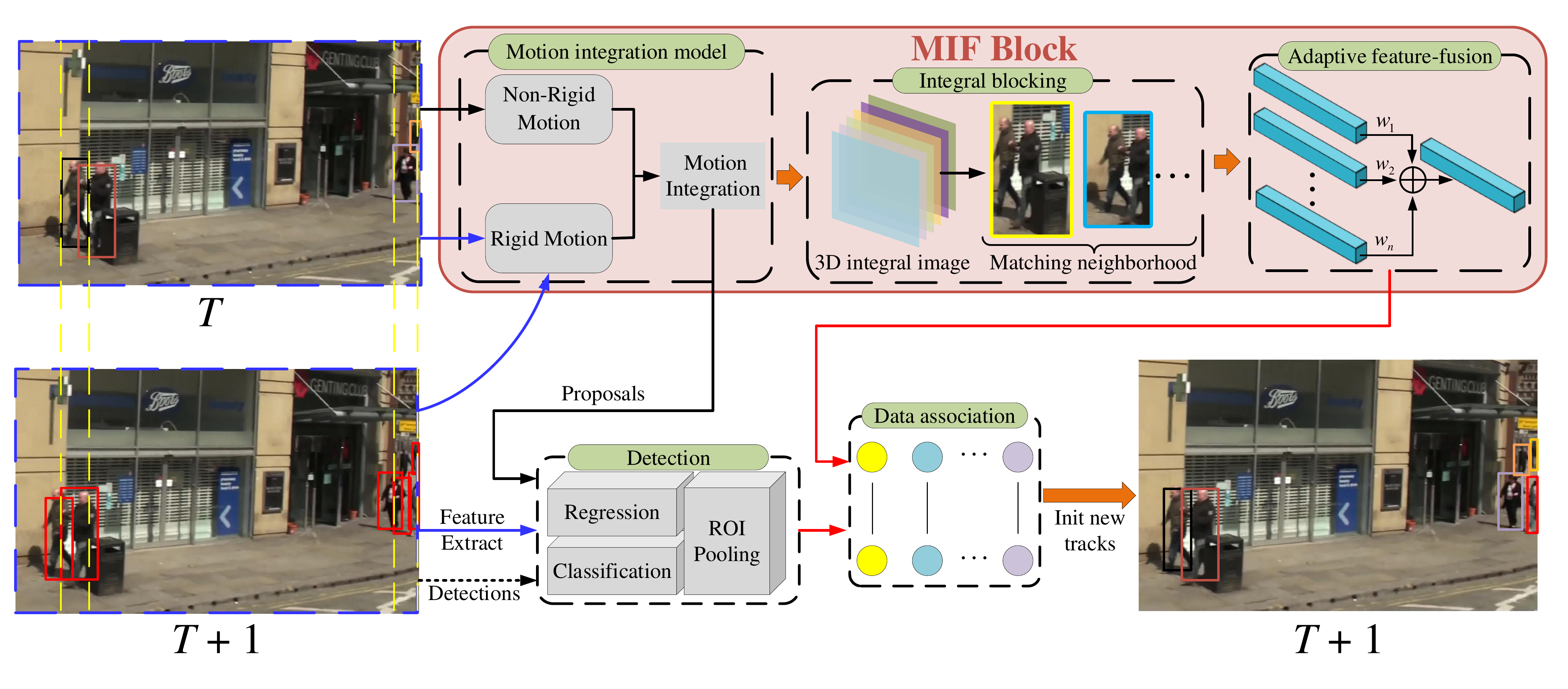}
  \caption{Illustration of the MIF architecture with an regression based tracker. For a given frame {\em T}, the integrated motions are applied to predict each track's positions in frame {\em T}+1 considering camera pose variances. Second, each track object will be constrained to a local searching region using 3D integral image and the trajectories' historical features are adaptively weighted for different detections. After the regression and classification of tracked boxes, new detections will be associated with the trajectories.}
  \label{framework}
\end{figure*}

\section{Proposed Method}
In this work, we propose a general architecture MIF for multi-object tracking. It can also be extended to the detection task. Our framework (See Fig.\ref{framework}) contains \textbf{M}otion integration considering non-rigid motion (pedestrian motion) and rigid motion (camera motion), the spatial blocking using the 3D \textbf{I}ntegral image and adaptive appearance feature \textbf{F}usion for pose alignment between detections and tracks. Besides, the spatial blocking module aims to apply spatial constraints to each tracked boxes, which is very time saving for both metric computation and graph construction.

We use Tracktor~\cite{Bergmann_2019_ICCV} as our baseline tracker, which treats tracking predictions and the provided detections as self-defined proposals to replace the region proposal network. Then proposals are passed to ROI Pooling block for regression and classification.

\subsection{Motion Integration}

In some cases, the IOU (intersection-over-union) based data association can outperform many general methods. This becomes possible due to the high quality detections and high frame rates. However, if there exists large pedestrian motion, camera motion or low frame rates, we must take them into consideration. As for camera motion, the pixel correspondence among sequential frames can be established by epipolar geometry (Ego Motion) constraints or affine transformation. With the assumption that the targets have slow motion and static shape, targets' states can be formulated as an optimization problem~\cite{wang2019exploit} with Ego motion, where $F$ is the fundamental matrix, and $x$ denotes the coordinates of target bounding boxes.
\begin{align}
\label{epipolar}
    \begin{split}
        f\left( {{x_{i,t + 1}}} \right) = \sum\limits_{i = 1}^4 {{{\left\| {x_{i,t{\rm{ + }}1}^TF{x_{i,t}}} \right\|}^2}}+ \left\| {\left( {{x_{3,t + 1}} - {x_{1,t + 1}}} \right) - \left( {{x_{3,t}} - {x_{1,t}}} \right)} \right\|_2^2
    \end{split}
\end{align}

In this method, the fundamental matrix needs to be estimated by feature matching without camera information. But feature points usually rely on the regions that contain lots of gradient information, which are also seriously interfered by human parts. As a consequence, the predicted targets' states will not be reliable. Here we study in combining rigid motion and non-rigid motion tightly using Enhanced Correlation Coefficient Maximization (ECC) and Kalman Filter.\\

Considering the space consistency, the pedestrian motion model needs to be processed before the camera motion model. In details, each target's position needs to be predicted by Kalman Filter firstly and aligned by the ECC model, which is named as Kalman+ECC. Besides, we have applied the fading memory to Kalman Filter to focus more on recent motions due to the uncertainty caused by camera motion and non-uniform motion states. Thus the Kalman+ECC motion model can be established as below:

\begin{equation}
    \left\{ \begin{array}{l}
    {s_{t + 1}} = warp\left(F{s_t}\right)\\
    {P_{t + 1}} = {\alpha}F{P_t}{F^T} + Q
    \end{array} \right.
    \label{motion model}
\end{equation}

Where $\alpha$ denotes the fading memory coefficient, $Q$ denotes the process covariance, $s$ and $P$ denote the predicted states and prior covariance of Kalman Filter, $warp$ denotes the ECC model.

However, the independent motion processing of Kalman+ECC solution will raise compatible problem. Thus we mix the camera motion and pedestrian motion together by using the affine matrix to adjust the integrated motion model. In this way, the integrated motion model can adapt to various motion scenes without pre-defined parameters. First, we define the intension of camera motion as Eq.\ref{intension}.
\begin{align}
Ic = 1-\frac{{\vec W \times \vec R}}{{{{\left\| {\vec W} \right\|}_2} \times {{\left\| {\vec R} \right\|}_2}}},R = \left[ {I;O} \right]
\label{intension}
\end{align}

Where the $Ic$ denotes the intension of camera motion. $\vec W$ denotes the vectorization of the affine matrix. The $R$ means the the affine matrix of static frames. $I$ is the identity matrix and $O$ is the all-zero matrix.

With the intension defined above, we can adjust the Kalman Filter by changing the state transition matrix:

\begin{equation}
\left\{ {\begin{array}{*{20}{l}}
{{s_{t + 1}} = warp\left( {{F_c}{s_t}} \right)}\\
{{P_{t + 1}} = \alpha {F_c}{P_t}{F_c}^T + Q}
\end{array}} \right.,{F_c} = \left[ {\begin{array}{*{20}{c}}
I&{\left( {dt + Ic} \right)I}\\
O&I
\end{array}} \right]
\label{motion integration}
\end{equation}
Where the $F_c$ denotes the adjusted state-transition matrix and the $dt$ means the original time step of Kalman Filter.

\subsection{Spatial Blocking via 3D Integral Image}
\label{integral blocking}
\begin{figure}[t]
  \centering
  \includegraphics[width=120mm]{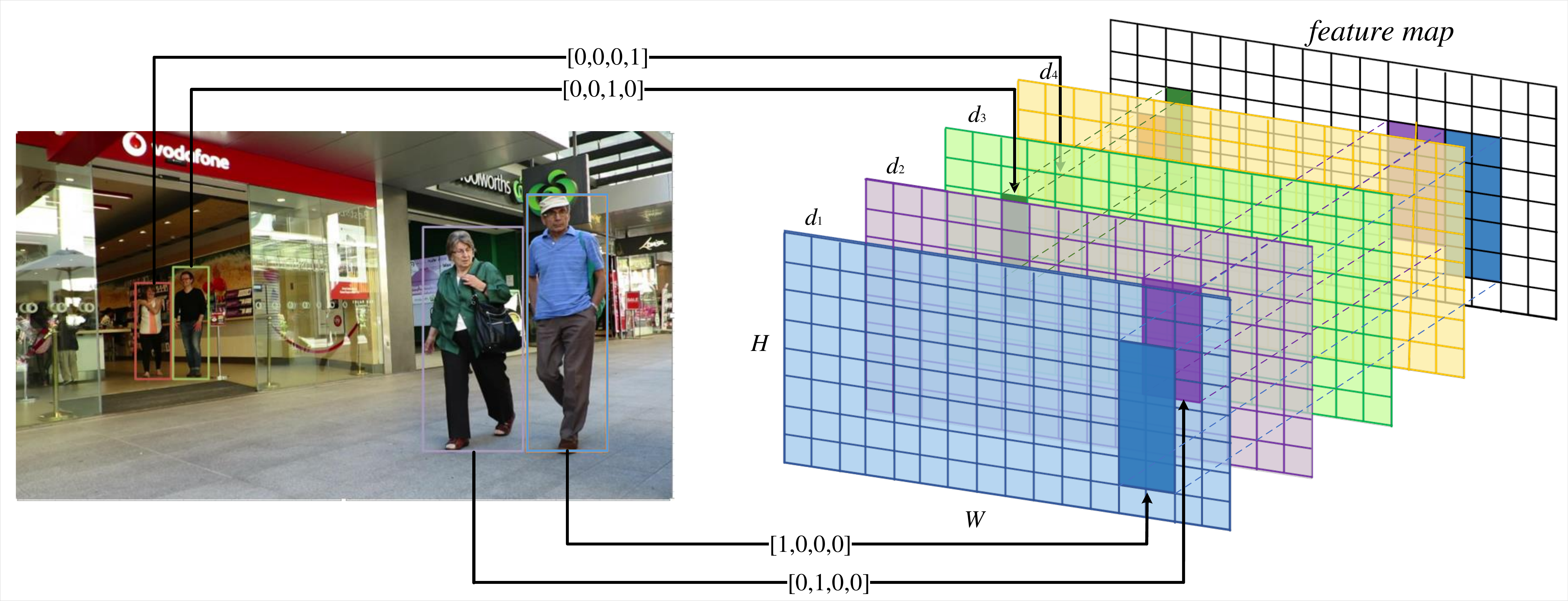}
  \caption{Illustration of feature mapping. The input image of $W\times H$ is divided into $M\times N$ cells to reduce space complexity. Thus each cell has the size of $\frac{W}{M} \times \frac{H}{N}$. Then each detection bounding box $d_i$ will be mapped into corresponding regions in the graph. Every cell contains a D-dim vector which uses one-hot encoding style, where the D denotes the number of detections in this frame. For example, [1 0 0 0] in cell$\left( {m,n} \right)$ represents that this cell overlaps with the first detection box.}
  \label{integral}
\end{figure}
\label{integral image}
The time complexity of calculating cost matrix between track bounding boxes and candidates is commonly $O\left( {{n^2}} \right)$. To assign each track bounding box with detections nearby, we transfer the detections into mask-based one-hot encoding descriptors (See Fig.\ref{integral}). This feature representation can be computed very rapidly using 3D integral image.

After getting the $M\times N$ cells of feature map, each cell contains the candidates' position information with D-dim vector. The 3D integral image at location $m,n$ contains the sum of candidates from $\left( {0,0} \right)$ to $\left( {m,n} \right)$ in feature map, including:
\begin{equation}
  I\left( {m,n} \right) = \sum\limits_{m' \le m,n' \le n} {f\left( {m',n'} \right)}
\end{equation}

Where $I\left( {m,n} \right)$ is the 3D integral image and ${f\left( {m',n'} \right)}$ is the feature map. We can simplify the process by dynamic programming:
\begin{align}\label{compute intergral}
    \begin{split}
         I\left( {m,n} \right) = I\left( {m,n - 1} \right) + I\left( {m - 1,n} \right) - I\left( {m - 1,n - 1} \right) + f\left( {m,n} \right)
    \end{split}
\end{align}

For each coming trajectory's bounding box $[{x_1},{x_2},{y_1},{y_2}]$, a spatial blocking region which contains several cells will be assigned. Using the 3D integral image, we can directly get the candidate lists of each spatial blocking region with constant complexity.
\begin{align}
    \begin{split}
         I\left( {{x_1}:{x_2},{y_1}:{y_2}} \right) &= I\left( {{x_2},{y_2}} \right) + I\left( {{x_1} - 1,{y_1} - 1} \right) \\
         &- I\left( {{x_1} - 1,{y_2}} \right) - I\left( {{x_2},{y_1} - 1} \right)
    \end{split}
\end{align}

Although the time complexity of data association is still $\mathcal O\left( {mn} \right)$. $m$ and $n$ denote the number of detections and trajectories. Most of the operations are assignment, addition and subtraction. Thus the time costs of this stage is actually reduced. Besides, it also requires space to save the 3D integral image, thus the space complexity is increased from $\mathcal O\left( {1} \right)$ to $\mathcal O\left( {n} \right)$.

\subsection{Adaptive Appearance Feature Fusion}
\begin{figure}[t]
    \centering
    \includegraphics[width=120mm]{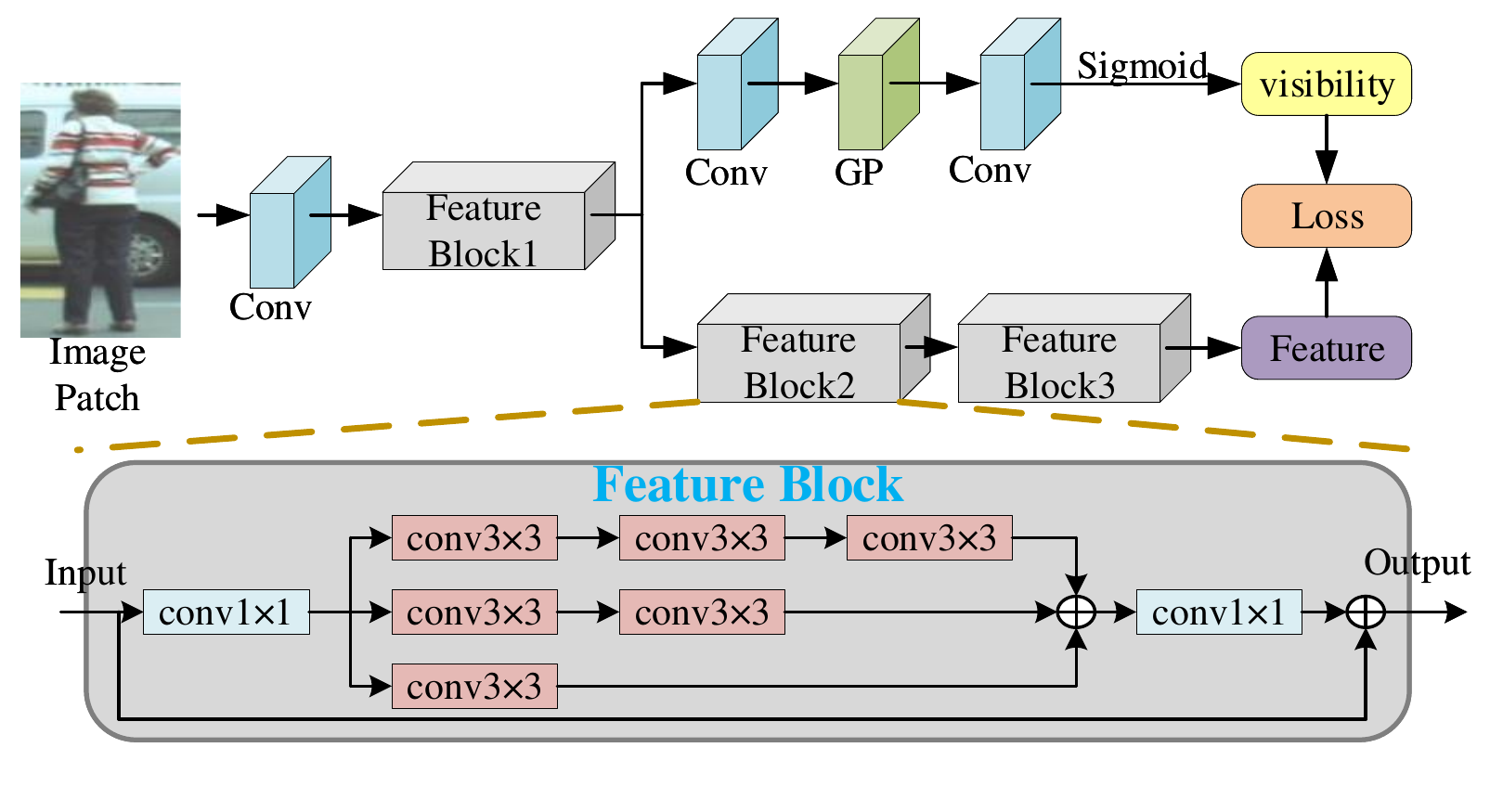}
      \caption{Architecture of our proposed occlusion aware appearance model. The appearance branch contains three feature blocks, which are formed as multi-scale residual inception blocks. The visibility prediction branch is performed after the first feature block of appearance branch.}
  \label{reid}
\end{figure}
\begin{figure}[t]
    \centering
    \includegraphics[width=100mm]{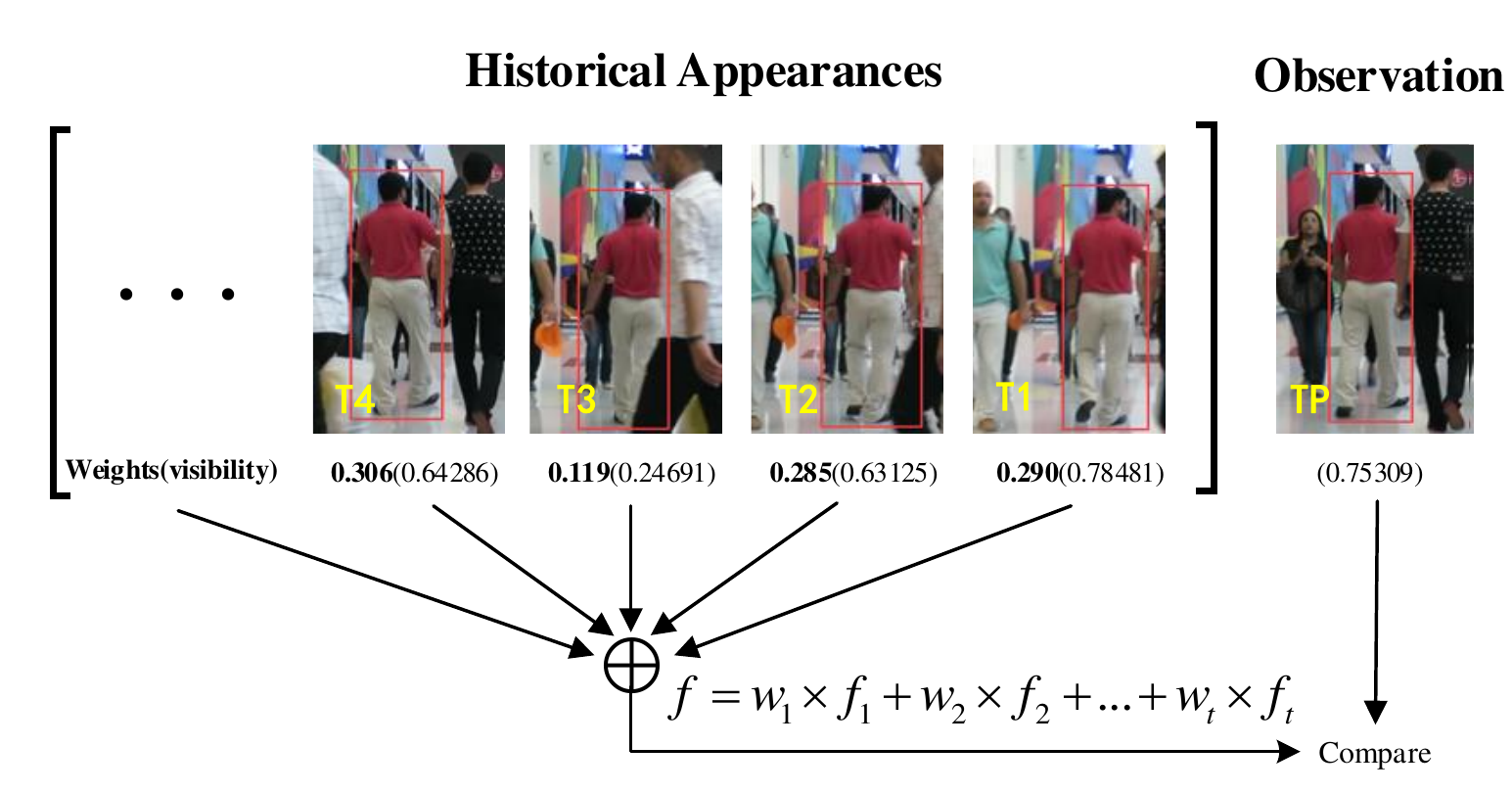}
      \caption{Illustration of adaptive feature fusion algorithm. Each historical appearance in the trajectory will be automatic weighted as the bold texts show. Besides, the visibility of each appearance is also shown within brackets.}
  \label{feature fusion}
\end{figure}

Considering the occlusion, different pose and scale of pedestrian, we propose a pose and occlusion aware adaptive appearance feature fusion model. This model consists of two aspects, including occlusion aware appearance model as shown in  Fig.~\ref{reid} and adaptive feature fusion as shown in Fig.~\ref{feature fusion}.

Resolution-invariant representation~\cite{DBLP:conf/ijcai/MaoZY19} has been proposed to address the scale and resolution misalignments in the field of person re-identification. Thus we proposed a light-weight feature block which is formed as a multi-scale residual inception block using cascade convolution to obtain different scales of receptive fields.
\begin{equation}
    {L_a} = \frac{1}{N}\sum\limits_{i = 1}^N {y_i^*log\left( {{y_i}} \right) + \left( {1 - y_i^*} \right)log\left( {1 - {y_i}} \right)}
    \label{crossentropy}
\end{equation}

The cross-entropy loss for the appearance model is defined as Eq.~\ref{crossentropy}. In contrast, inspired by the position sensitive mask~\cite{chen2019aggregate}, we trained the visibility prediction branch combined with the appearance model. However, most objects are totally visible, which brings an imbalanced problem. The multi-task loss is designed by adding a coefficient $\phi$ to leverage the imbalance between different visibility of objects.

\begin{equation}
    Loss = {L_a} + \frac{\phi }{N}\sum\limits_{i = 1}^N {{{\left( {{v_i} - v_i^*} \right)}^2}}
\end{equation}

With the appearance features, the similarity measurement between trajectories and candidates can be simply performed as a feature selection or feature fusion question except for those aggregate end-to-end frameworks. However, feature selection carries great misalignment risks. Thus we propose an adaptive feature fusion model combined with visibility, scale, aspect and time information. As for the differences between a candidate's feature with one trajectory's historical features in those four aspects, we simply use min-max normalization to uniform the dimensions. Also, those four aspects are weighted to sum up to get the total distance as shown on Fig.~\ref{feature fusion}.
\begin{equation}
    {d} = {\lambda _1}{d_{scale}} + {\lambda _2}{d_{aspect}} + {\lambda _3}{d_{visibility}} + {\lambda _4}{d_{time}}
\end{equation}

The weight coefficients of each trajectory's historical features can be calculated as:
\begin{equation}
    weight{s_i} = \frac{{\exp \left( { - {d_i}} \right)}}{{\sum\nolimits_j {\exp \left( { - {d_j}} \right)} }}
\end{equation}

\begin{algorithm}[t]
%    \SetAlgoNoLine  %去掉之前的竖线
 \caption{MIF based tracking algorithm}
  \KwIn{Video sequences $\mathcal I = \left\{ {{{\mathcal I}_1},{{\mathcal I}_2}, \cdots ,{{\mathcal I}_T}}\right\}$ and provided detections $\mathcal D = \left\{ {{D_1},{D_2}, \cdots ,{D_T}} \right\}$.}
  \KwOut{Trajectories $\mathcal T$}
  $\mathcal T \leftarrow \phi $\;
  ${\mathcal L}_k$: Lost length of  ${\mathcal T}_k$\;
  ${\mathcal F}_k$: Appearance Feature of ${\mathcal T}_k$\;
  \For{$t = 1, \cdots, T$}{
        Extract features of ${I_t}$\;
        Apply Integrated motions to ${\mathcal T}_k$ referring to Eq.~\ref{motion integration}\;

        ${B_t},{S_t} \leftarrow Regress\_and\_Classify\left\{ {{D_t},{\mathcal T}_k} \right\}$\;
        $\mathcal B \leftarrow NMS\left( {[{B_t},{D_t}], thresh} \right)$\;
        Build 3D integral image with $\mathcal B$ as in Section ~\ref{integral image}\;
        Extract appearance features of each $\mathcal B$\;
        \For{${{\mathcal T}_k} \in \mathcal T$}{
            $b \leftarrow SpatialBlock\left( {B,{{\mathcal T}_k}} \right)$\;
            ${F_k} \leftarrow AdaptiveWeightedFeatures\left( {b,{{\mathcal T}_k}} \right)$\;
            $Cost \leftarrow GetCost\left( {b,F_k,{{\mathcal T}_k}} \right)$ by Eq.~\ref{cost}\;
            }
        Associate the $\mathcal B$ with $\mathcal T$ using $Cost$\;
        \For{${{\mathcal T}_k} \in \mathcal T$}{
            \eIf{Assigned with $B_i$}{
                ${\mathcal T}_k \leftarrow B_i$\;
                ${\mathcal F}_k \leftarrow {\mathcal F}_k + F_i$\;
                ${\mathcal L}_k = 0$;
            }
            {
            \eIf{${\mathcal L}_k >$ time gap}{
                $\mathcal T = \mathcal T - {\mathcal T}_k$\;
                }
                {
                    ${\mathcal L}_k ++$\;
                }
                }
        }
        $\mathcal T \leftarrow \mathcal T + \left\{ {\mathcal B - {{\mathcal T}_k}} \right\}$\;
    }
    Delete the inactive trajectories.
 \label{MIFT}
\end{algorithm}

\subsection{MIF based Tracker}

Combined with the proposed MIF method mentioned above, we can easily extend it to track multiple objects with an existing tracker. Here we use Mahalanobis distance to evaluate motion distance with the covariances of Kalman Filter.
\begin{equation}
  {d_m} = {\left( {det - track} \right)^T}{S^{ - 1}}\left( {det - track} \right)
\end{equation}

Where the $S$ denotes the system uncertainty of Kalman Filter. Then the appearance distances are computed by normalized cosine metric. Actually, the motion distance is more suitable for the short-term association and appearance distance is more likely used for long-term association. Thus we proposed a balanced way to integrate both of them.
\begin{equation}
    \begin{split}
    w = miss\_rat{e^{time\_gap}}\\
    d = w{d_m} + \left( {1 - w} \right){d_a}
    \end{split}
  \label{cost}
\end{equation}

Where $time\_gap$ denotes the time length since the trajectories get lost. With the inactive trajectory's length increasing, the preserved position will be unreliable. Thus the appearance feature will be adopted for compensation and the weights of appearance cost will be increased along with the $time\_gap$.

\begin{figure*}[t]
\setlength{\abovedisplayskip}{3pt}
  \centering
  \includegraphics[width=120mm]{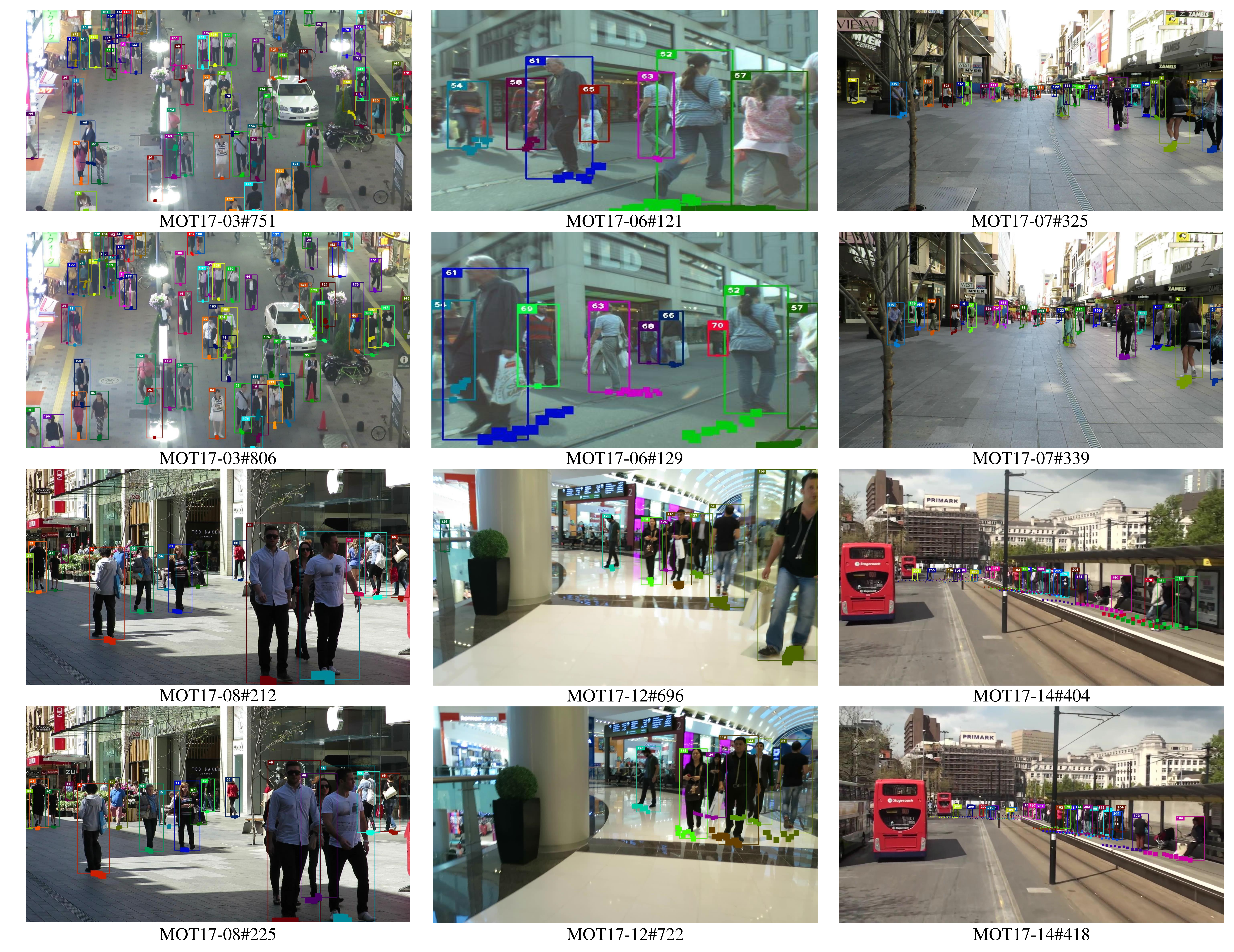}
  \caption{Visualization of tracking results of MOT17 test sets with MIFT using SDP detector.}
  \label{visualization}
\end{figure*}

\section{Experiments}

\subsection{Experiment Setup}
\label{experiment}
Experiments are conducted on the widely used multi-object tracking benchmark MOTChallenge\footnote{https://motchallenge.net/}. This benchmark consists of several challenging pedestrian tracking and detection sequences with frequent occlusion and crowded scenes, which vary in the angle of camera views, objects scales and frame rates. Our MIF based tracker has been evaluated on three separate challenge, named 2D MOT2015~\cite{MOTChallenge2015}, MOT16 and MOT17~\cite{MOT16}. In addition, both of the MOT16 and MOT17 contain the same 7 train sequences and 7 test sequences. The difference is that the MOT17 benchmark provides three different public detections (DPM~\cite{felzenszwalb2009object}, Faster R-CNN~\cite{ren2015faster}, SDP~\cite{yang2016exploit})  with increasing performance, but the MOT16 benchmark only provides the DPM one. The 2D MOT2015 benchmark also provides ACF~\cite{dollar2014fast} detections for 22 sequences.

\textbf{Evaluation Metric}. Evaluation is carried out according to the widely accepted CLEAR MOT metrics~\cite{bernardin2008evaluating}, including accuracy of the multiple object tracking (MOTA), the number of ID Switches (ID Sw.), the total number of false positive (FP) and false negative (FN) et al. Among this metrics, the MOTA and ID Sw. can quantify the main two aspects as object coverage and identify.

\textbf{Implementation Details}. The proposed approach is implemented in Pytorch and runs on a desktop with a CPU of 10 cores@2.2GHz and two RTX2080Ti GPUs. The fading memory and time step of Kalman Filter are set to 1.2 and 0.15. To leverage the efficiency and accuracy, the 3D integral image are divided into 16 $\times$ 8 blocks. As for the appearance model, the network is trained for 150 epochs with the learning rate of 3e-3 and the batch size is 64. The input image patch size for the appearance model is 64 $\times$ 256 and the dimensions of feature is 512. Due to variable number of targets, the feature extraction stage will cost much time. Experiments show that with the batchsize increasing, the speed of feature extraction grows linearly at first. Then it tends to be stable. Thus, all of the targets' features will be extracted with a fixed batchsize (26) whether the number of features is divisible by the batchsize. Also we re-implement the Faster RCNN detector by changing the anchor aspect to $\{1.0, 2.0, 3.0\}$ with the pre-trained weights trained on COCO datasets. Note that both the object detector and re-id model are trained on scratch with a multi-scale strategy. Specifically, the reconnection mechanism of trajectories is only applied to the scenes with camera motion in which the time gaps can reach 10. Each trajectory can hold at most 26 historical features. During the post-process stage, the trajectories whose length less than 5 are removed.

\subsection{Ablation Study}

The ablation study was evaluated on the validation sets which are extracted from MOT17 train sets. Since we have used the parts of train sets for training, here we only use the validation sets for ablation study.

\begin{table}[t]
\begin{minipage}{0.48\linewidth}
   \caption{Ablation study in terms of different motion models. The Ego denotes the Epipolar Geometry model and MI denotes the integrated motion model.}
   \label{ablation1}%
   \centering
    \begin{tabular}{cccc}
    \hline\noalign{\smallskip}
    Method & MOTA$\uparrow$   & IDF1$\uparrow$   & ID Sw.$\downarrow$  \\
    \noalign{\smallskip}
    \hline
    \noalign{\smallskip}
    Ego   & 53.31 & 44.22 & 1829 \\
    ECC   & 59.2  & 59.44 & \textbf{481} \\
    Kalman & 59.33 & 58.81 & 604 \\
    Kalman+ECC & 59.48 & 59.86 & 569 \\
    \noalign{\smallskip}
    \hline
    \noalign{\smallskip}
    MI & \textbf{60.23} & \textbf{59.87} & 509 \\
    \hline
    \end{tabular}%
\end{minipage}
\begin{minipage}{0.48\linewidth}
   \caption{Comparisons of different appearance models. The first and the latest selected historical features are compared with features fused in average way and our proposed adaptive way. }
   \label{ablation2}%
   \centering
    \begin{tabular}{cccc}
    \hline\noalign{\smallskip}
    Method & MOTA$\uparrow$  & IDF1$\uparrow$  & ID Sw.$\downarrow$  \\
    \noalign{\smallskip}
    \hline
    \noalign{\smallskip}

    ReID(avg)  & 57.48  & 53.15  & 1486  \\
    ReID(latest)  & 57.76  & 53.38  & 1123 \\
    ReID(fusion)   & 57.92  & 53.78  & 1238  \\

    \noalign{\smallskip}
    \hline
    \noalign{\smallskip}
    ReID+MI      & \textbf{60.38}  & \textbf{61.47}  & \textbf{484} \\
    \hline
    \end{tabular}%
\end{minipage}
\end{table}%

\textbf{Motion Integration}. The different motion models' influences on the MOT task are shown in Table~\ref{ablation1}. Here the Ego model (Epipolar Geometry) referring to Eq.~\ref{epipolar} is also added to the experiments. And the MI denotes the motion integration referring to Eq.~\ref{motion integration}. The baseline model is our re-implemented Tractor++~\cite{Bergmann_2019_ICCV}. It can be seen from the Table~\ref{ablation1} that using both of the Kalman Filter and ECC models without integration (Kalman+ECC) referring to Eq.~\ref{motion model} can be better than just adopting any one of them. Moreover, the integrated motion model referrring to Eq.~\ref{motion integration} shows obvious advantages on MOTA and IDF1 to the non-integrated Kalman+ECC model. Also it's obvious that the epipolar model (Ego) can not handle the motion alignments well. Notice that the ID Sw. of single ECC model is slightly smaller than the integrated motion model. The fact is that the number of trajectories tracked by the integrated motion model is much more than the ECC model.

\begin{figure}[tb]
\centering
\begin{minipage}[t]{0.48\textwidth}
\centering
\includegraphics[width=\columnwidth, height=40mm]{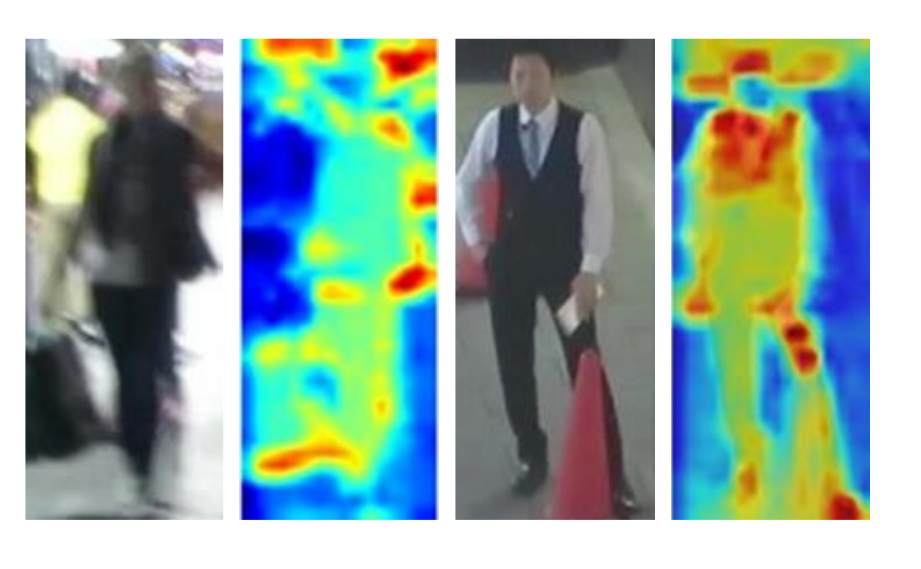}
\caption{Heatmaps' visualizations of the first feature block's outputs.}
\label{heatmap}
\end{minipage}
\begin{minipage}[t]{0.48\textwidth}
\centering
\includegraphics[width=\columnwidth, height=40mm]{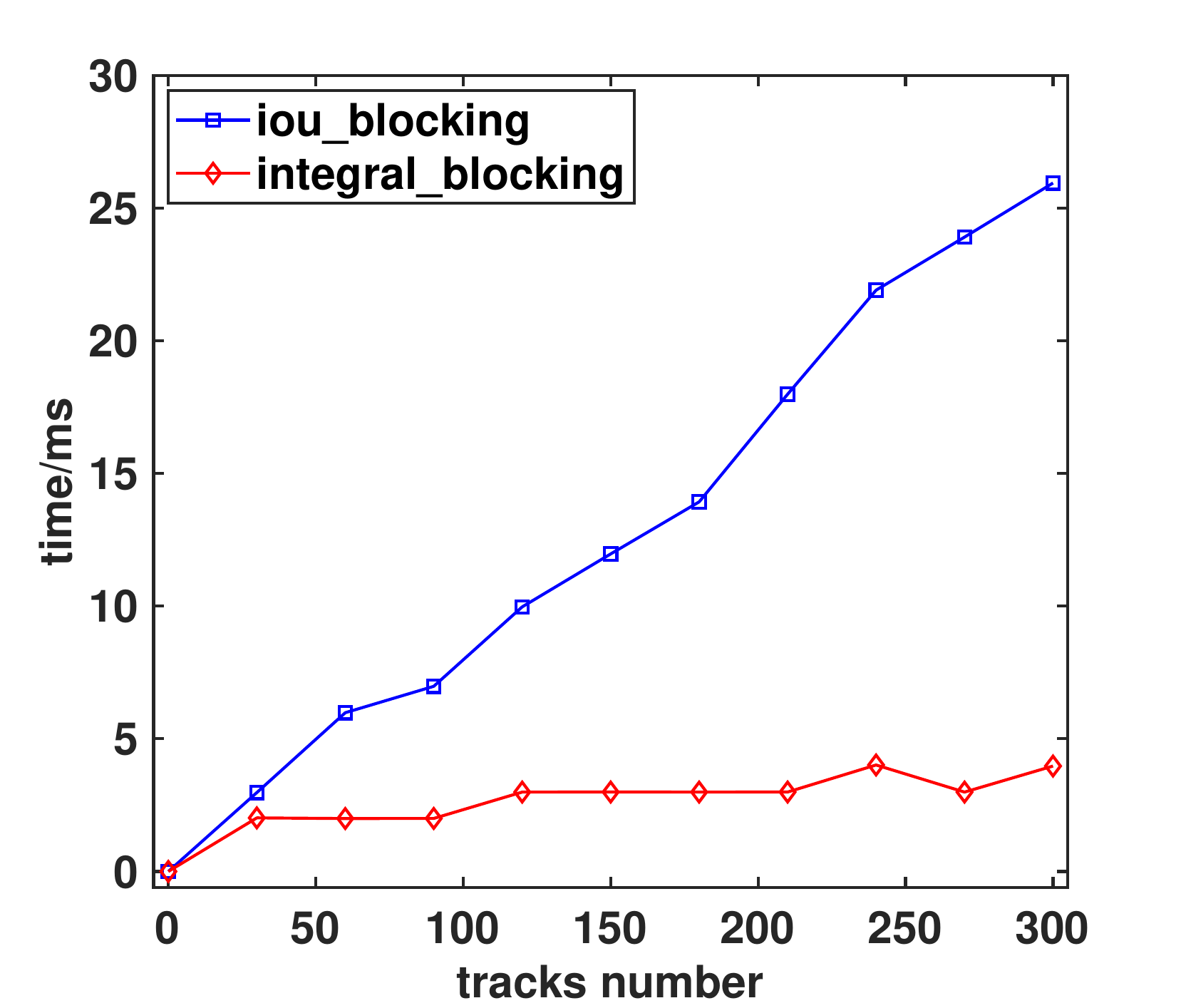}
\caption{Speed comparison using IOU based blocking and 3D integral image based blocking}
\label{iou_vs_integral}
\end{minipage}
\end{figure}

\textbf{Appearance Model}. In order to evaluate the appearance model and adaptive feature fusion methods, several different feature selection and fusion methods are evaluated in Table~\ref{ablation2}. The ReID denotes our proposed occlusion aware appearance model. In Table~\ref{ablation2}, we experimentally select the representative historical features by simply using the latest feature in trajectories and fuse the features by averaging them and our adaptive feature fusion model. In contrast, the adaptive appearance feature fusion model can improve the MOTA and IDF1 scores much better. Combining the short and long cues, the MOTA score is improved by 1.5\%,2.46\% and IDF1 is improved by 1.6\%,7.69\% than simply using the motion integration model and adaptive feature fusion model. Moreover, the visualizations of appearance features are showed in Fig.~\ref{heatmap}. Since we trained the appearance model with the visibility prediction branch, the model focuses more on the foreground targets.

\textbf{3D Integral image}. In order to demonstrate the speed of our proposed spatial blocking method using 3D integral image named as integral\_blocking, we also present an IOU based region blocking method named as iou\_blocking for comparison. If the detections have overlap with the extended region of track's bounding boxes, then the detections will be assigned to the tracks. The speed comparison is shown in Fig.~\ref{iou_vs_integral}, which shows the significant advantage of our integral\_blocking in speed to the IOU based one, especially when there exists a large number of trajectories or the detections per frame.

\begin{table}[t]

\caption{Comparison of our method with the methods on the MOT Challenge. \textbf{O} denotes the online methods.}
  \centering
  \resizebox{\linewidth}{!}{
  \scriptsize
    \begin{tabular}{lccccccccc}
    \toprule
    Methods  & MOTA$\uparrow$ & IDF1$\uparrow$ & MT$\uparrow$   & ML$\downarrow$   & FP$\downarrow$   & FN$\downarrow$   & ID Sw.$\downarrow$ & Frag$\downarrow$ & Hz$\uparrow$ \\
    \midrule
    \multicolumn{10}{c}{2D MOT 2015} \\
    \midrule
    \textbf{Ours(O)}  & 46.7  & 51.6  & 29.4\% & 25.7\% & 11003 & \textbf{20839} & 878   & 1265  & 6.7  \\
    MPNTrack~\cite{braso2019learning} & \textbf{48.3} & \textbf{56.5} & \textbf{32.2\%} & \textbf{24.3\%} & 9640  & 21629 & 504   & 1074  & 9.3  \\
    Tracktor\textbf{(O)}~\cite{Bergmann_2019_ICCV} & 44.1  & 46.7  & 18.0\% & 26.2\% & 6477  & 26577 & 1318  & 1790  & 0.9  \\
    KCF\textbf{(O)}~\cite{chu2019online}   & 38.9  & 44.5  & 16.6\% & 31.5\% & 7321  & 29501 & 720   & 1440  & 0.3  \\
    AP\_HWDPL\_p\textbf{(O)}~\cite{chen2017online} & 38.5  & 47.1  & 8.7\% & 37.4\% & \textbf{4005} & 33203 & 586   & 1263  & 6.7  \\
    STRN\textbf{(O)}~\cite{Xu_2019_ICCV}  & 38.1  & 46.6  & 11.5\% & 33.4\% & 5451  & 31571 & 1033  & 2665  & \textbf{13.8 } \\
    AMIR\textbf{(O)}~\cite{sadeghian2017tracking}  & 37.6  & 46.0  & 15.8\% & 26.8\% & 7933  & 29397 & 1026  & 2024  & 1.9  \\
    JointMC~\cite{keuper2018motion} & 35.6  & 45.1  & 23.2\% & 39.3\% & 10580 & 28508 & \textbf{457} & \textbf{969} & 0.6  \\
    \midrule
    \multicolumn{10}{c}{MOT16} \\
    \midrule
    \textbf{Ours(O)}  & \textbf{60.1} & 56.9  & \textbf{26.1\%} & \textbf{29.1\%} & 6964  & \textbf{65044} & 739   & 951   & 6.9 \\
    MPNTrack~\cite{braso2019learning} & 55.9  & \textbf{59.9} & 26.0\% & 35.6\% & 7086  & 72902 & 431   & 921   & 11.9 \\
    Tracktor\textbf{(O)}~\cite{Bergmann_2019_ICCV} & 54.4  & 52.5  & 19.0\% & 36.9\% & \textbf{3280} & 79149 & 682   & 1480  & 1.5 \\
    NOTA~\cite{chen2019aggregate}  & 49.8  & 55.3  & 17.9\% & 37.7\% & 7248  & 83614 & 614   & 1372  & \textbf{19.2} \\
    HCC~\cite{ma2018customized}   & 49.3  & 50.7  & 17.8\% & 39.9\% & 5333  & 86795 & \textbf{391} & \textbf{535} & 0.8  \\
    LSSTO\textbf{(O)}~\cite{feng2019multi} & 49.2  & 56.5  & 13.4\% & 41.4\% & 7187  & 84875 & 606   & 2497  & 2.0  \\
    TNT~\cite{wang2019exploit}   & 49.2  & 56.1  & 17.3\% & 40.3\% & 8400  & 83702 & 606   & 882   & 0.7  \\
    AFN~\cite{shen2018tracklet}   & 49.0    & 48.2  & 19.1\% & 35.7\% & 9508  & 82506 & 899   & 1383  & 0.6  \\
    \midrule
    \multicolumn{10}{c}{MOT17 } \\
    \midrule
    \textbf{Ours(O)}  & \textbf{60.1} & 56.4  & \textbf{28.5\%} & \textbf{28.1\%} & 23168 & \textbf{199483} & 2556  & 3182  & \textbf{7.2} \\
    MPNTrack~\cite{braso2019learning} & 55.7  & 59.1  & 27.2\% & 34.4\% & 25013 & 223531 & 1433  & 3122  & 4.2 \\
    LSST~\cite{feng2019multi}  & 54.7  & \textbf{62.3} & 20.4\% & 40.1\% & 26091 & 228434 & \textbf{1243} & 3726  & 1.5 \\
    Tracktor\textbf{(O)}~\cite{Bergmann_2019_ICCV} & 53.5  & 52.3  & 19.5\% & 36.6\% & \textbf{12201} & 248047 & 2072  & 4611  & 1.5 \\
    LSSTO\textbf{(O)}~\cite{feng2019multi} & 52.7  & 57.9  & 17.9\% & 36.6\% & 22512 & 241936 & 2167  & 7443  & 1.8 \\
    JBNOT~\cite{henschel2019multiple} & 52.6  & 50.8  & 19.7\% & 35.8\% & 31572 & 232659 & 3050  & 3792  & 5.4 \\
    FAMNet~\cite{Chu_2019_ICCV} & 52.0  & 48.7  & 19.1\% & 33.4\% & 14138 & 253616 & 3072  & 5318  & 0.0  \\
    TNT~\cite{wang2019exploit}   & 51.9  & 58.1  & 23.1\% & 35.5\% & 36164 & 232783 & 2288  & \textbf{3071} & 0.7  \\
    \bottomrule
    \end{tabular}%
    }

    \label{benchmark}
\end{table}%

\subsection{Evaluation on Benchmarks}

The performance of our MIF based tracker (MIFT) has been evaluated on all of the MOT test sequences\footnote{ISE\_MOT17R in https://motchallenge.net/results/MOT17/}. The results officially published are shown in the Table~\ref{benchmark}. Both the online and batch methods are demonstrated in the same table. As shown in the table, our tracker (MIFT) outperforms all of the existing online trackers on most of the metrics, especially for MOTA, IDF1, MT, ML, FN. Also, our online tracker has much lower computational costs compared to most of the trackers. On the 2DMOT2015 challenge, due to the poor quality of detections, our proposed tracker performs slightly worse than the only one batch method (MPNTracker). To summarize, our proposed method can significantly improve the ML and MOTA due to the motion integration, which keeps the trajectories continuity. With the 3D integral image, the speed of our tracker is much faster than the baseline tracker (Tracktor).

Other qualitative results are shown on Fig.~\ref{visualization}. All of the test sequences are tracked by MIFT with SDP detections as observations. Obviously, our proposed tracker can get precise tracked boxes of targets. Also, the MIFT is robust to the irregular camera motions(such as MOT17-06 and MOT17-14),crowded scenes(such as MOT17-03), different camera viewpoints(such as MOT17-03 and MOT17-07). Especially on the MOT17-14 sequences, which are captured by a fast moving camera that is mounted on bus in a busy intersection. Our proposed tracker can still be able to track targets in a stable and persistent way.

\begin{table}[t]

\caption{Comparison with the state-of-arts MOT Detection methods. Our MIFD detector is a MIF based detector, which is combined with a re-implemented Faster RCNN detector with FPN.}
  \centering

    \begin{tabular}{cccccc}
    \toprule
    Method & AP$\uparrow$    & MODA$\uparrow$  & FAF$\downarrow$   & Precision$\uparrow$ & Recall$\uparrow$ \\
    \midrule
    MSCNN~\cite{cai2016unified} & \textbf{0.89}  & \textbf{76.7}  & 2.8   & 86.2  & 91.3 \\
    POI~\cite{yu2016poi}   & 0.89  & 67.1  & 4.8   & 78.7  & 92.1 \\
    ViPeD~\cite{amato2019learning} & 0.89  & -14.4 & 20.8  & 46.4  & \textbf{93.2} \\
    FRCNN~\cite{ren2015faster} & 0.72  & 68.5  & \textbf{1.7}   & \textbf{89.8}  & 77.3 \\
    \midrule
    \textbf{FRCNN+FPN} & 0.88  & 65.9  & 5.1   & 77.7  & 92.4 \\
    \textbf{MIFD}  & \textbf{0.88}  & \textbf{67.4}  & \textbf{4.9}   & \textbf{78.6}  & \textbf{92.6} \\
    \bottomrule
    \end{tabular}%

  \label{motdet}%
\end{table}%

\subsection{Extension to Detection}
In this section, we first re-implement the Faster RCNN detector with FPN. Then the new detector will be adopted to replace the public detections in the tracking task. The results shown in the Table~\ref{motdet} demonstrate that the MIF based detecter (MIFD) can also obtain promising results on MOT17 Det challenge. Moreover, the MIF based detector outperforms the non-MIF based detector (FRCNN+FPN) on almost every metric.

\section{Conclusion}
In this paper, we explore the refinements in motion and appearance model. Thus a general architecture named MIF (\textbf{M}otion integration, 3D \textbf{I}ntegral image and adaptive appearance feature \textbf{F}usion) is proposed, which can be embedded into both tracking and detection tasks. Experiments are conducted on a widely used MOT Challenge, which demonstrate the advantages of both of our MIF based tracker (MIFT) and MIF based detector (MIFD). Specifically, since the motion and appearance models are commonly used in tracking methods, our proposed methods can help overcome their interaction effects and misalignments. Moreover, the association between detections and trajectories can be simplified via our proposed 3D integral image, which is significantly efficient as shown in Figure~\ref{iou_vs_integral} and the last column of Table~\ref{benchmark}.

\clearpage

\bibliographystyle{splncs04}
\bibliography{MIFT}
\end{document}